\documentclass[conference]{IEEEtran}
\IEEEoverridecommandlockouts
\usepackage{cite}
\usepackage{amsmath,amssymb,amsfonts}
\usepackage{algorithmic}
\usepackage{graphicx}
\usepackage{textcomp}
\usepackage{xcolor}
\def\BibTeX{{\rm B\kern-.05em{\sc i\kern-.025em b}\kern-.08em
    T\kern-.1667em\lower.7ex\hbox{E}\kern-.125emX}}

\usepackage{tabularx}               
\usepackage{multirow}               
\newcolumntype{Y}{>{\centering\arraybackslash}X}   
\usepackage{stackengine}
\newcommand\xrowht[2][0]{\addstackgap[.5\dimexpr#2\relax]{\vphantom{#1}}}
\usepackage{array}
\newcolumntype{L}[1]{>{\raggedright\let\newline\\\arraybackslash\hspace{0pt}}m{#1}}
\newcolumntype{C}[1]{>{\centering\let\newline\\\arraybackslash\hspace{0pt}}m{#1}}
\newcolumntype{R}[1]{>{\raggedleft\let\newline\\\arraybackslash\hspace{0pt}}m{#1}} 
\newcolumntype{Z}[1]{>{\minwd c{#1}}c<{\endminwd}}
\def\minwd#1#2#3\endminwd{\stackengine{0pt}{#3}{\rule{#2}{0pt}}{O}{#1}{F}{F}{L}}
\newcolumntype{Q}{@{}c@{}}
\usepackage{colortbl}
\usepackage{color,xcolor}
\definecolor{hsuhhred}{rgb}{0.7725, 0.0, 0.2588}
\definecolor{hsuhhred1}{rgb}{0.6471, 0.0, 0.2039}
\definecolor{hsuhhred2}{rgb}{0.7294, 0.2392, 0.3882}
\definecolor{hsuhhred3}{rgb}{0.8118, 0.4902, 0.5804}
\definecolor{hsuhhred4}{rgb}{0.8902, 0.7294, 0.7804}
\usepackage{hhline}
\usepackage{subcaption}
\usepackage{multirow}
\usepackage{booktabs}

\makeatletter 
\newcommand{\linebreakand}{%
\end{@IEEEauthorhalign}
\hfill\mbox{}\par
\mbox{}\hfill\begin{@IEEEauthorhalign}
}
\makeatother 

\begin{document}

\title{Drone Detection and Tracking with YOLO and a Rule-based Method}

\author{\IEEEauthorblockN{Purbaditya Bhattacharya}
\IEEEauthorblockA{\textit{Dept. of Signal Proc. and Comm.} \\
	\textit{Helmut Schmidt University}\\
	Hamburg, Germany \\
	bhattacp@hsu-hh.de}
\and
\IEEEauthorblockN{Patrick Nowak}
\IEEEauthorblockA{\textit{Dept. of Signal Proc. and Comm.} \\
	\textit{Helmut Schmidt University}\\
	Hamburg, Germany \\
	patrick.nowak@hsu-hh.de}
}

\maketitle

\begin{abstract}
Drones or unmanned aerial vehicles are traditionally used for military missions, warfare, and espionage. However, the usage of drones has significantly increased due to multiple industrial applications involving security and inspection, transportation, research purposes, and recreational drone flying. Such an increased volume of drone activity in public spaces requires regulatory actions for purposes of privacy protection and safety. Hence, detection of illegal drone activities such as boundary encroachment becomes a necessity. Such detection tasks are usually automated and performed by deep learning models which are trained on annotated image datasets. This paper builds on a previous work and extends an already published open source dataset. A description and analysis of the entire dataset is provided. The dataset is used to train the YOLOv7 deep learning model and some of its minor variants and the results are provided. Since the detection models are based on a single image input, a simple cross-correlation based tracker is used to reduce detection drops and improve tracking performance in videos. Finally, the entire drone detection system is summarized.
\end{abstract}


\section{Introduction}
\label{sec:Introduction}

Drones or unmanned aerial vehicles (UAV) are primarily used in military operations, industrial and commercial applications. But recreational drone flying has also increased in urban areas which can be concerning for public safety and privacy. Hence, regulation in terms of legal zones and other requirements are usually put in place. At the same time, it is necessary to monitor if any violation of such regulations have taken place. An automated drone detection system can therefore generate an alert based on any illegal activity and enable a capturing system to act. Such detection systems can operate based on data processed by different sensors. In this context, infrared and visual cameras are used as two of many sensors which might provide better visibility under different environmental conditions.

\begin{figure}[htpb]
	\centering
	\includegraphics[width=0.98\columnwidth]{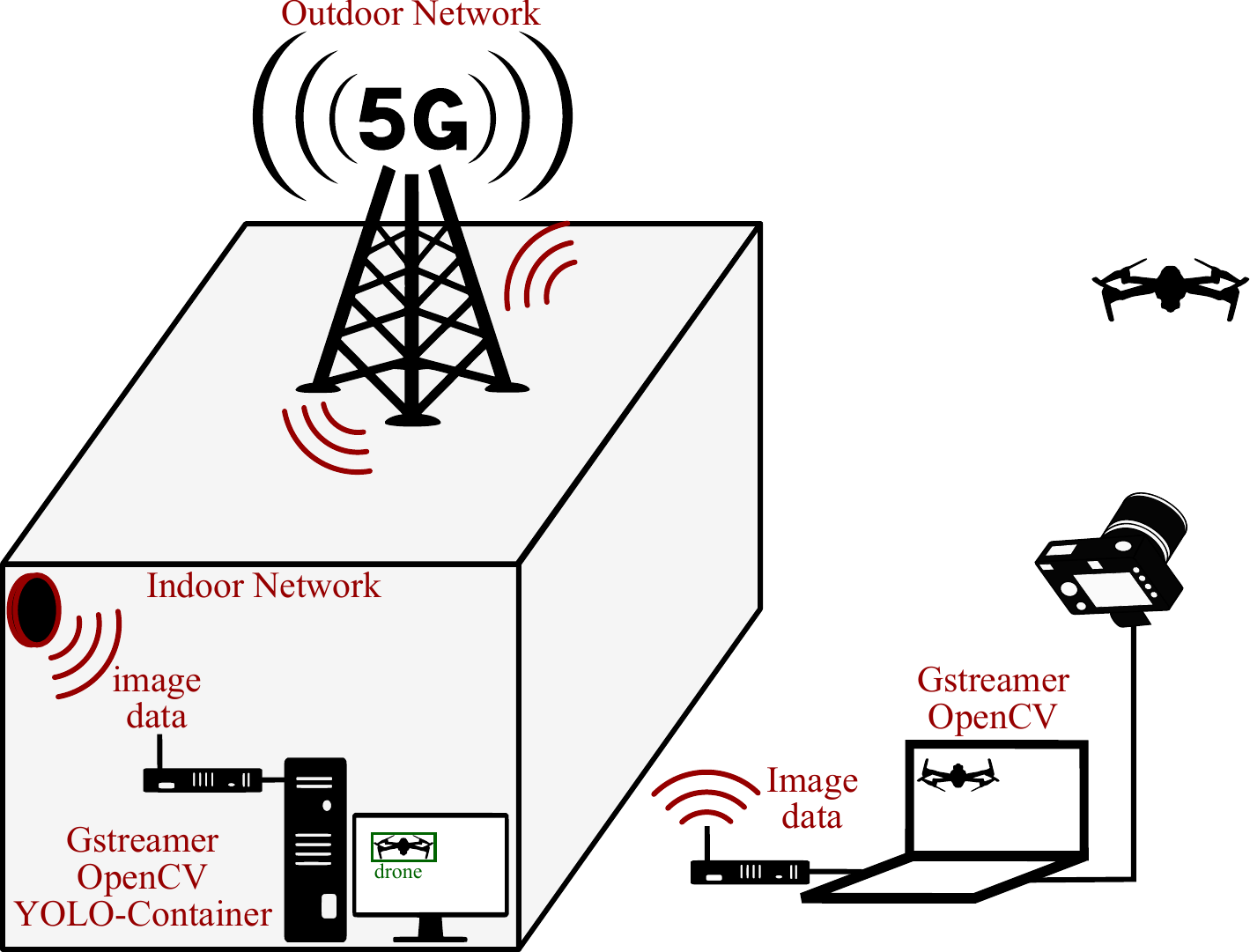}
	\caption{Overview of the drone detection system.}
	\label{fig:systemoverview}
\end{figure}

A part of this work builds on our previous work \cite{bhattacharyaDronedataset} on a published dataset of infrared images. The previous dataset is a collection of annotated infrared images captured by a FLIR and an InfraTec Camera at two distinct locations - the university campus and one of the city harbours. In this work, a set of annotated color images, also captured in the same locations, are added to the existing dataset. Additionally, more videos are recorded at the harbour location with additional backgrounds and the corresponding frames are annotated and added to the dataset.

With the incremental improvement in deep learning based methods, they are primarily used for multiple object detection tasks \cite{FasterRCNN, SSD, efficientDet, bhattacharyaLiveshore}. More commercial multi-sensor or vision systems now use a trained and properly scaled deep learning model for improved object detection. YOLO is one of the frontrunners in object detection and is widely used for these tasks. Hence, such models are used for drone detection \cite{dtecBand, liu2024realtimedetectionsmalluavs}. The larger drone dataset is created with the objective of developing and training the YOLO convolutional neural network (CNN) models and integrate them in a drone detection system. The base model and models modified by the addition of smaller modules are trained and compared. 

The following sections provide an overview of the drone detection system followed by an analysis of the datasets. A brief description of the modified YOLO models are provided in the next section along with an evaluation of their performance. Finally, a simple ensemble method for improving the detection performance and tracking is introduced followed by a conclusion.

\section{System Overview}
\label{sec:Overview} 

In this section, the system overview as illustrated in Fig.\,\ref{fig:systemoverview} is described. The camera placed in the campus area covered by 5G outdoor network records a flying drone and the processed video stream is transmitted to the indoor network. The transmission is then received by the computer which runs the drone detection algorithm. After detection, the computer can display the results or transmit it further to another machine. In order to send or recieve videos, the GStreamer multimedia application framework \cite{gstreamer} is used. The constructed GStreamer pipelines primarily include data transmission and reception via UDP, processing and conversion, encoding and decoding, and buffering. The received stream is processed by a customized pre-trained YOLOv7 model in a Docker \cite{merkel2014docker} container which has Python OpenCV built with GStreamer support. Finally, the result is displayed.

\section{Dataset}
\label{sec:Dataset} 

\begin{figure}
	\centering
	\includegraphics[width=0.98\columnwidth]{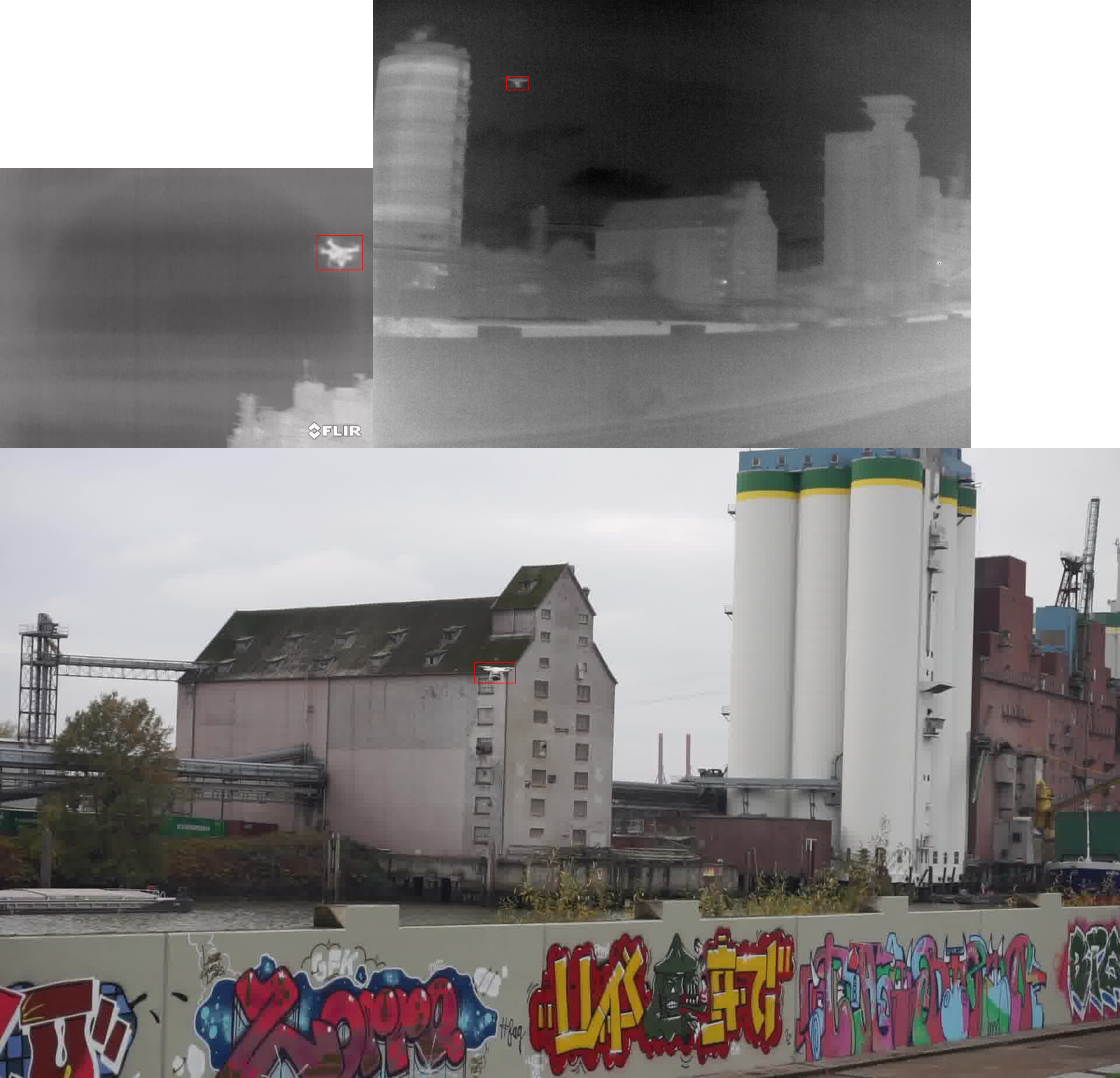}
	\caption{Comparison of the image resolution of all cameras: FLIR Scion OTM366 (top left - $640 \times 480$ pixels), InfraTec VarioCAM HD Z (top right - $1024 \times 768$ pixels), and Sony $\alpha$6000 (bottom - $1920 \times 1080$ pixels)}
	\label{fig:compareCameras}
\end{figure}

To create the dataset, videos of flying drones are recorded with three different cameras, followed by the extraction of frames and their annotation by various methods. The following subsections provide a brief description of the process.

\begin{figure}[htbp]
	\centering
	\includegraphics[width=0.98\columnwidth]{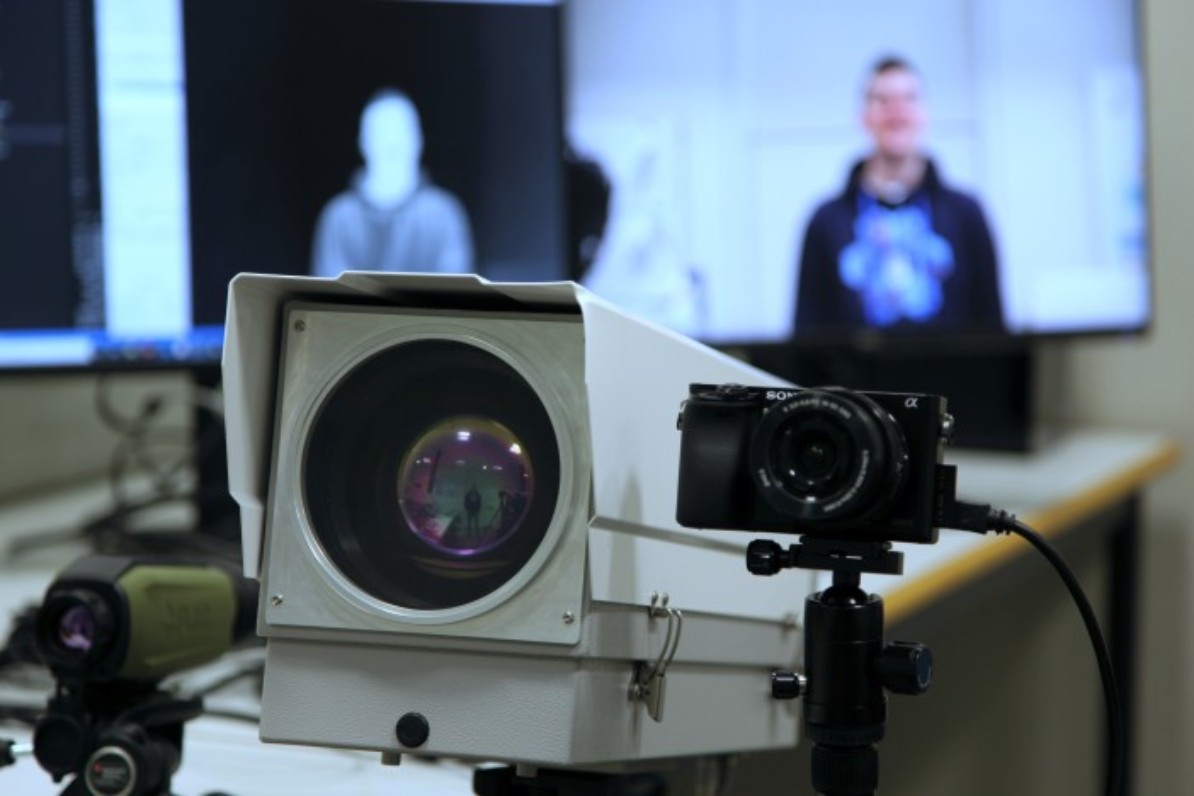}
	\caption{Used cameras: FLIR Scion OTM366 (left), InfraTec VarioCAM HD Z (center), and Sony $\alpha$6000 (right).}
	\label{fig:cameras}
\end{figure}

\subsection{Video Recording}
The original dataset is constructed from drone videos captured at the university football field and the Wilhelmsburg harbour using a FLIR Scion OTM366 infrared camera and an InfraTec VarioCAM HD Z equipped with a zoom lens (25-150\,mm) with image resolutions of $640 \times 480$ pixels and $1024 \times 768$ pixels, respectively. The color videos are recorded at the aforementioned locations with a Sony $\alpha$6000 camera with an image resolution of $1920 \times 1080$ pixels. The initial videos are recorded independently with each cameras from multiple perspectives with some videos being recorded on different days.

\begin{figure*}
	\centering
	\begin{subfigure}[b]{0.32\textwidth}
		\centering
		\includegraphics[scale=0.12]{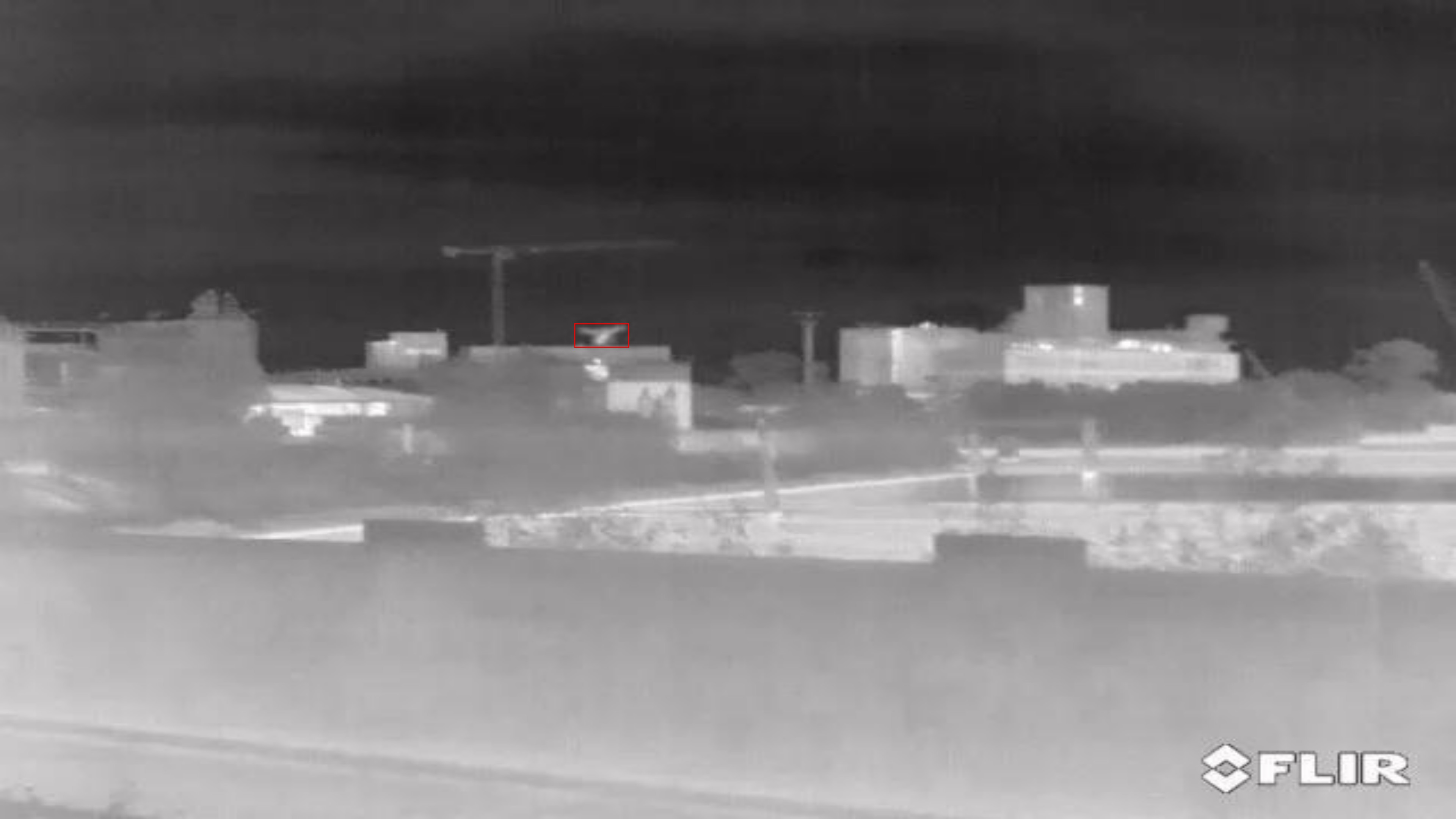}
		\caption{FLIR-Harbour}
		\label{fig:FLIRHafen}
	\end{subfigure}
	\hfill
		\begin{subfigure}[b]{0.32\textwidth}
		\centering
		\includegraphics[scale=0.12]{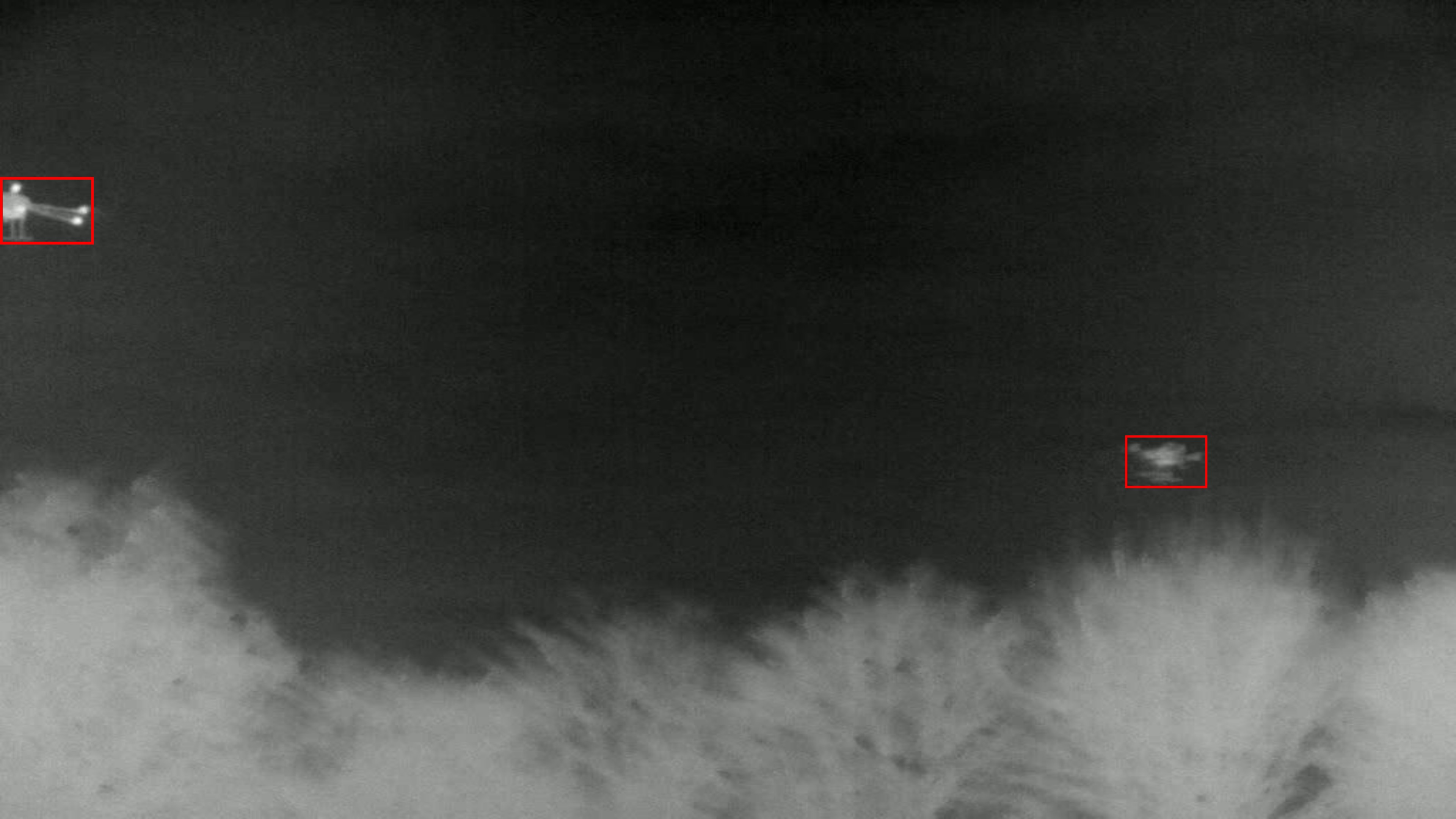}
		\caption{InfraTec-Campus}
		\label{fig:InfraTecHSU}
	\end{subfigure}
	\hfill
	\begin{subfigure}[b]{0.32\textwidth}
		\centering
		\includegraphics[scale=0.12]{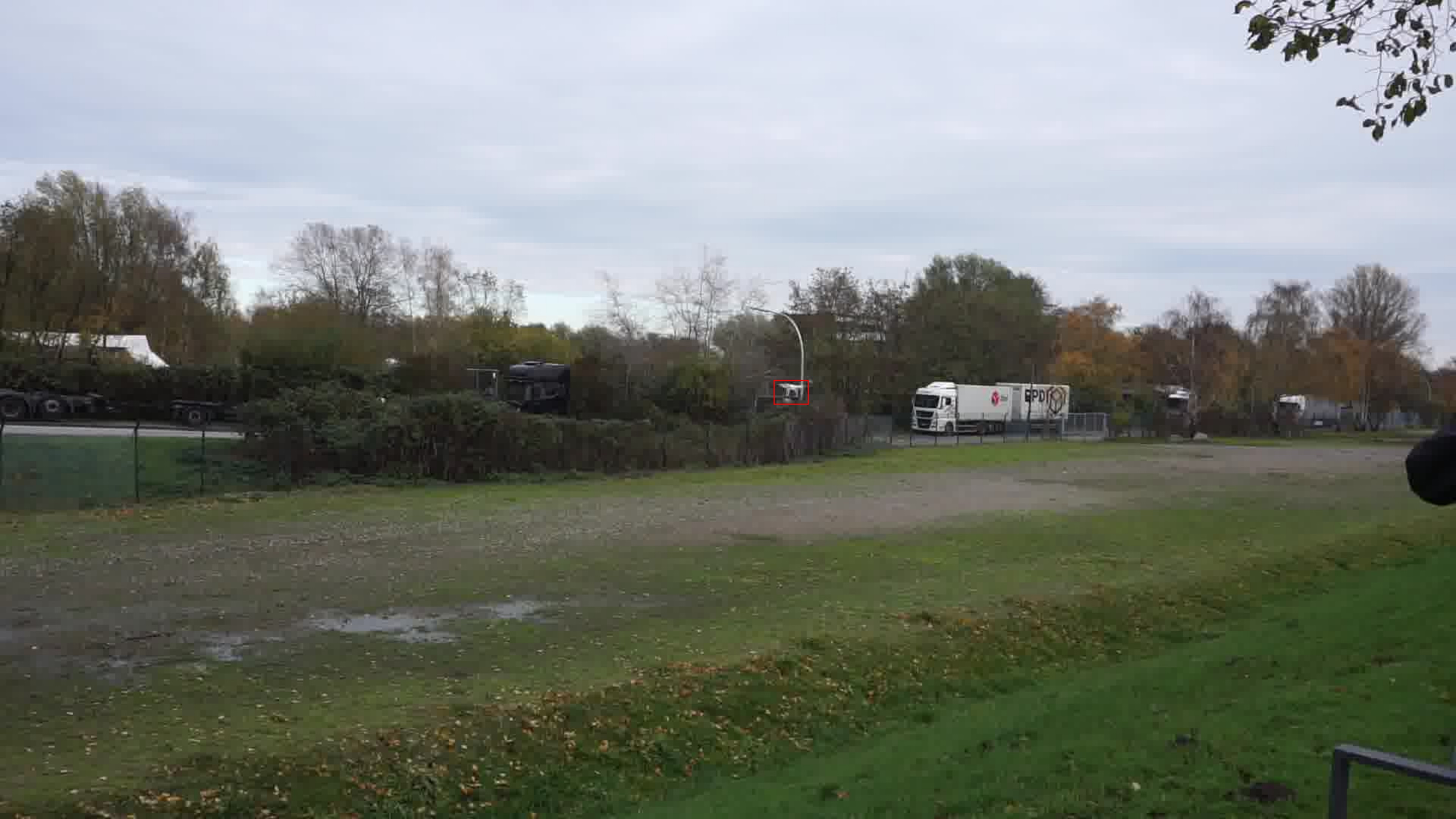}
		\caption{Sony-Harbour}
		\label{fig:SonyHafen1}
	\end{subfigure} \\ \vspace{0.25cm}
	\begin{subfigure}[b]{0.32\textwidth}
		\centering
		\includegraphics[scale=0.12]{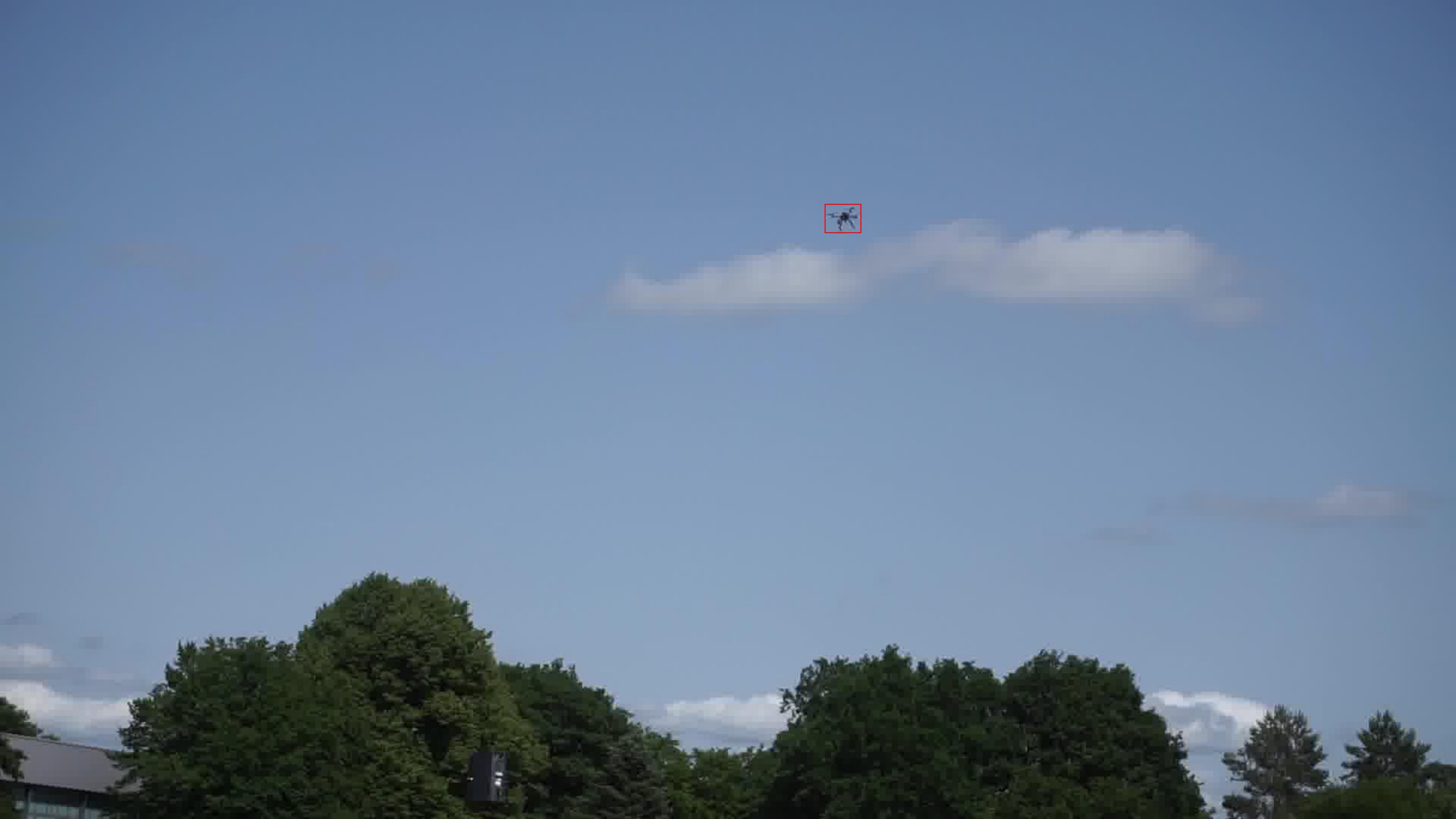}
		\caption{Sony-Campus}
		\label{fig:SonyHSU1}
	\end{subfigure}
	\hfill
	\begin{subfigure}[b]{0.32\textwidth}
		\centering
		\includegraphics[scale=0.12]{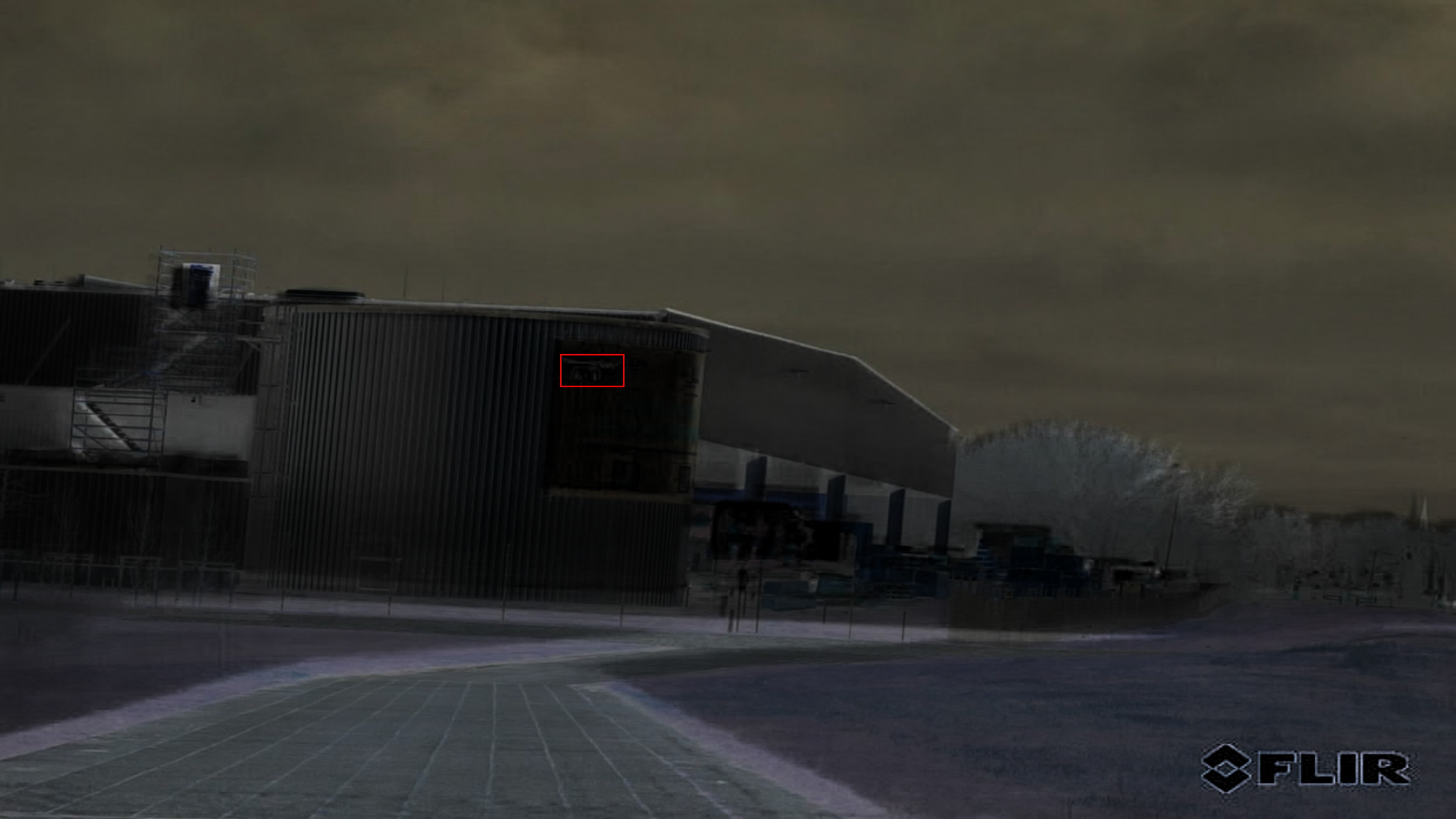}
		\caption{Multi-channel (Sony, FLIR)-Harbour}
		\label{fig:SonyFLIRHafen1}
	\end{subfigure}
	\hfill
	\begin{subfigure}[b]{0.32\textwidth}
		\centering
		\includegraphics[scale=0.12]{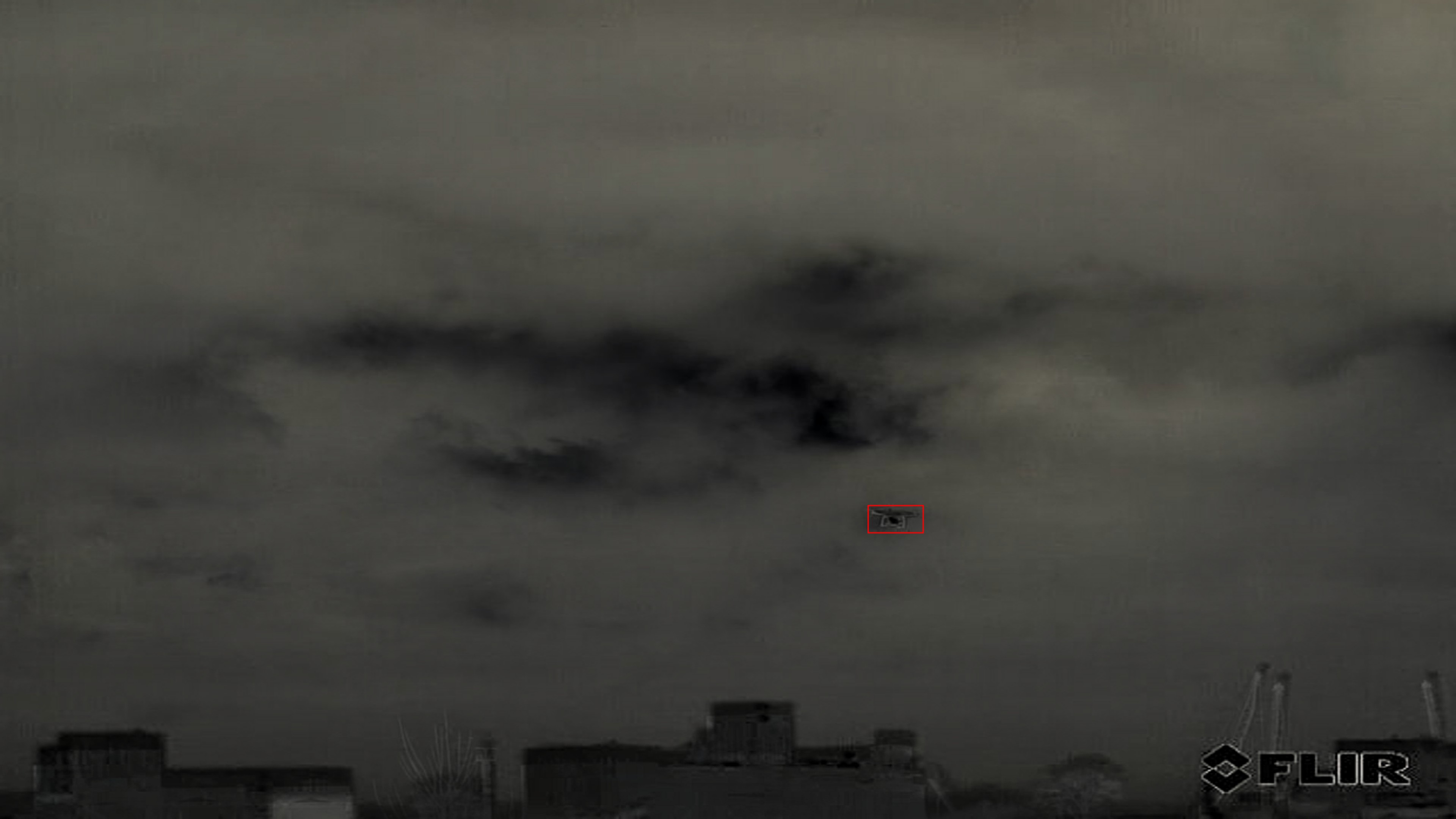}
		\caption{Multi-channel (Sony, FLIR)-Harbour}
		\label{fig:SonyFLIRHafen2}
	\end{subfigure}
	\caption{Exemplary images of the dataset recorded with FLIR Scion OTM366 at (\subref{fig:FLIRHafen}) harbour, with InfraTec VarioCAM HD Z at (\subref{fig:InfraTecHSU}) campus, with Sony $\alpha$6000 at (\subref{fig:SonyHafen1}) harbour and (\subref{fig:SonyHSU1}) campus, and with FLIR Scion OTM366 and Sony $\alpha$6000 at (\subref{fig:SonyFLIRHafen1}, \subref{fig:SonyFLIRHafen2}) harbour and combined.}
	\label{fig:exampleDataset}
\end{figure*}

Finally, all the cameras are mounted on a steerable tripod and additional videos are recorded at the Harbour. The new videos have more background information compared to the initial recordings and all cameras have very similar perspectives while recording the drone. Figure\,\ref{fig:compareCameras} shows some example images captured by the cameras. The camera setup on the tripod is shown in Fig.\,\ref{fig:cameras}. Example images from each location are also shown in Fig.\,{\ref{fig:exampleDataset}}

\subsection{Dataset Labelling}
\label{sec:Dataset Labelling}

The LabelImg \cite{tzutalin2015} graphical annotation tool is used to manually annotate the drones in the color images extracted from the video recordings. The annotations are done in the Pascal VOC \cite{everingham2016} format. As described in \cite{bhattacharyaDronedataset}, an automated annotation script based on a simple bounding box tracking method is used to accelerate the process followed by intermediate manual correction of any wrong detections. Drones that are difficult to spot are either not annotated or labelled as a difficult example. After the manual annotation of the initial infrared and color datasets the base YOLO model is used to train them. 

This trained model is used to detect drones in the second dateset constructed with the FLIR infrared and color images. Based on the detection results, initial annotations are generated and incorrect annotations are manually corrected. Since feature matching proved to be very challenging and inconsistent, automatic registration methods yielded poor results. Instead a transformation matrix is manually generated based on multiple trials, followed by a simpler cross-correlation based matching of small local regions around the drone. Image pairs are discarded when a registration yields poor results. Backgrounds in the image pairs are shifted with respect to each other in the multi-channel image but for a single class classification task, the impact of such a problem should be less.

\section{Dataset Analysis}
\label{sec:Dataset Analysis}

The previous dataset already consists of 66,438 images and 71,520 annotated drones. In 2,617 images there were no drones. In 10,130 cases, the annotated drone is marked as difficult because it is barely distinguishable from the background or mostly outside the image. The dataset prepared with infrared images is initially divided into normal and difficult images depending on the visibility of drones due to clutter, occlusion, or heavy blurring. In this work, the relatively easy images are selected where the drones are mostly visible by the naked eye. It is however noteworthy that the set of normal images do contain a substantial number of challenging images where the drones are partially occluded/ blurred or the contrast between the drones and background is not big. Additionally, the total number of final images used during training of a model is further reduced due to some images being blurry or the drones being very tiny.

The color image dataset contains a total of 37,508 images with 37,842 annotated drones. Instances where the drone is difficult to perceive are not annotated at all in this case. All of the images are used to train a model.

Finally, the additional dataset generated from the final recordings at the harbour consists of 40,059 infrared and color images each. However, this dataset is not used for training any model. Rather, an attempt is made to register the infrared and color images based on detections from a pre-trained model, so that the location of drones coincide relatively well and the images can be merged as 4-channel image data. Traditaional image processing and registration algorithms failed to reliably find the correct transformations and projections due to the difference in features between the infrared and color images as well as the quality of the images. The attempt to register the images manually is also made difficult due to imperfect synchronization and small relative movements of cameras on the tripod while following the drones. This has lead to background mismatches in most images. Table \ref{tab:dataset} provides an overview of all datasets.

\begin{table*}[bt]
	\centering
	\footnotesize
	\caption{Total number of images and annotated drones contained in the final dataset used for training separated into different types. Additionally, the number of drones marked as difficult is given.}
	\vspace{2mm}
	\label{tab:dataset}
	\begin{tabular}{|L{2.5cm}|Q|C{2cm}|Q|C{1cm}|C{1cm}|C{1cm}|C{1cm}|C{2cm}|Q|C{2cm}|} 
		\hhline{-~-~-----~-}
		\vspace{-0.3cm}\xrowht{0.4cm}\centering{\multirow{2}{*}{Cameras}} & \hspace{0.08cm}
		& \vspace{-0.3cm}\xrowht{0.4cm}\multirow{2}{*}{\begin{tabular}[c]{@{}c@{}}Number of\\ images\end{tabular}} & \hspace{0.08cm}
		& \multicolumn{4}{c|}{\xrowht{0.4cm}Number of drones per image}
		& \vspace{-0.3cm}\xrowht{0.4cm}\multirow{2}{*}{\begin{tabular}[c]{@{}c@{}}Number of\\ drones\end{tabular}} & \hspace{0.08cm}
		& \vspace{0.05cm}\xrowht{0.4cm}\multirow{2}{*}{\begin{tabular}[c]{@{}c@{}}Number of\\ drones marked\\ as difficult\end{tabular}}
		\\ \hhline{~~~~----~~~}
		\xrowht{0.4cm} & 
		& \xrowht{0.4cm} &
		& \xrowht{0.4cm}1
		& \xrowht{0.4cm}2
		& \xrowht{0.4cm}3
		& \xrowht{0.4cm}4
		& \xrowht{0.4cm} &
		&
		\\ \hhline{=~=~=====~=}
		\xrowht{0.3cm}FLIR & 
		& \xrowht{0.3cm}40,028 &
		& \xrowht{0.3cm}\phantom{0}33,038
		& \xrowht{0.3cm}\phantom{0}6,815
		& \xrowht{0.3cm}\phantom{0}175
		& \xrowht{0.3cm}\phantom{0}0
		& \xrowht{0.3cm}47,193 &
		& \xrowht{0.3cm}\phantom{0}6,290
		\\ \hhline{-~-~-----~-}
		\xrowht{0.3cm}InfraTec & 
		& \xrowht{0.3cm}23,793 &		
		& \xrowht{0.3cm}\phantom{0}23,259
		& \xrowht{0.3cm}\phantom{0}534
		& \xrowht{0.3cm}\phantom{0}0
		& \xrowht{0.3cm}\phantom{0}0
		& \xrowht{0.3cm}24,327 &
		& \xrowht{0.3cm}\phantom{0}3,840
		\\ \hhline{-~-~-----~-}
		\xrowht{0.3cm}\bf{Infrared (total)} & 
		& \xrowht{0.3cm}63,821 &
		& \xrowht{0.3cm}\phantom{0}56,297
		& \xrowht{0.3cm}\phantom{0}7,349
		& \xrowht{0.3cm}\phantom{0}175
		& \xrowht{0.3cm}\phantom{0}0
		& \xrowht{0.3cm}71,520 &
		& \xrowht{0.3cm}\phantom{0}10,130
		\\ \hhline{=~=~=====~=}
		\xrowht{0.3cm}Sony  & 
		& \xrowht{0.3cm}37,508 &
		& \xrowht{0.3cm}\phantom{0}37,208
		& \xrowht{0.3cm}\phantom{0}283
		& \xrowht{0.3cm}\phantom{0}0
		& \xrowht{0.3cm}\phantom{0}17
		& \xrowht{0.3cm}37,842 &
		& \xrowht{0.3cm}0
		\\ \hhline{-~-~-----~-}
		\xrowht{0.3cm}\bf{Color (total)} & 
		& \xrowht{0.3cm}37,508 &
		& \xrowht{0.3cm}\phantom{0}37,208
		& \xrowht{0.3cm}\phantom{0}283
		& \xrowht{0.3cm}\phantom{0}0
		& \xrowht{0.3cm}\phantom{0}17
		& \xrowht{0.3cm}37,842 &
		& \xrowht{0.3cm}0
		\\ \hhline{=~=~=====~=}
		\xrowht{0.3cm}\bf{Multichannel} & 
		& \xrowht{0.3cm}40,059 &
		& \xrowht{0.3cm}\phantom{0}40,059
		& \xrowht{0.3cm}\phantom{0}0
		& \xrowht{0.3cm}\phantom{0}0
		& \xrowht{0.3cm}\phantom{0}0
		& \xrowht{0.3cm}40,059 &
		& \xrowht{0.3cm}0
		\\ \hhline{-~-~-----~-}
	\end{tabular}
\end{table*}

\begin{figure}[htbp]
	\centering
	\hspace{-3em}
	\includegraphics[scale = 0.64]{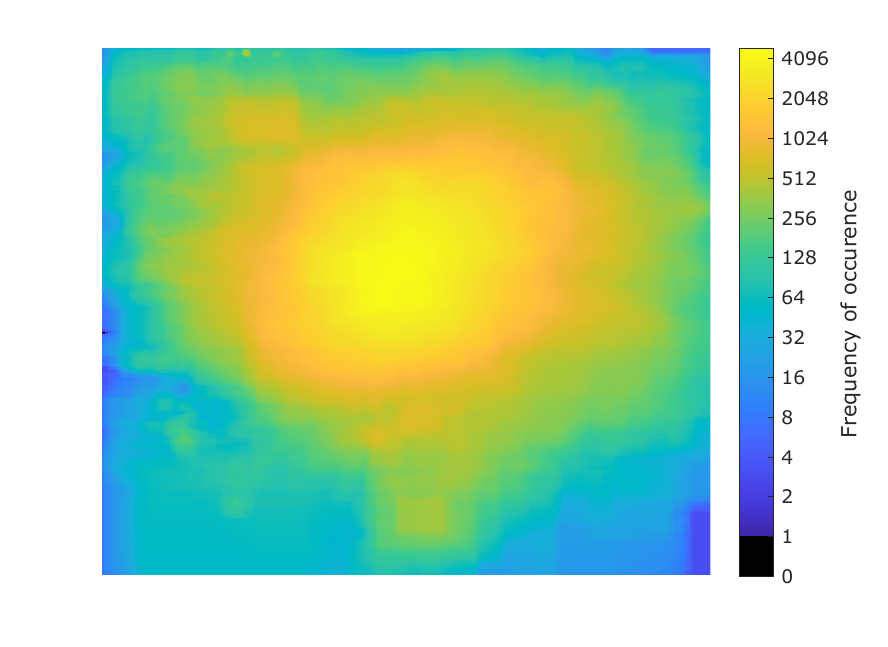}
	\caption{Annotation heatmap of the drones recorded by the Infrared cameras.}
	\label{fig:heatmap_infrared}
\end{figure}

\begin{figure}[htbp]
	\centering
	\hspace{-3em}
	\includegraphics[scale = 0.64]{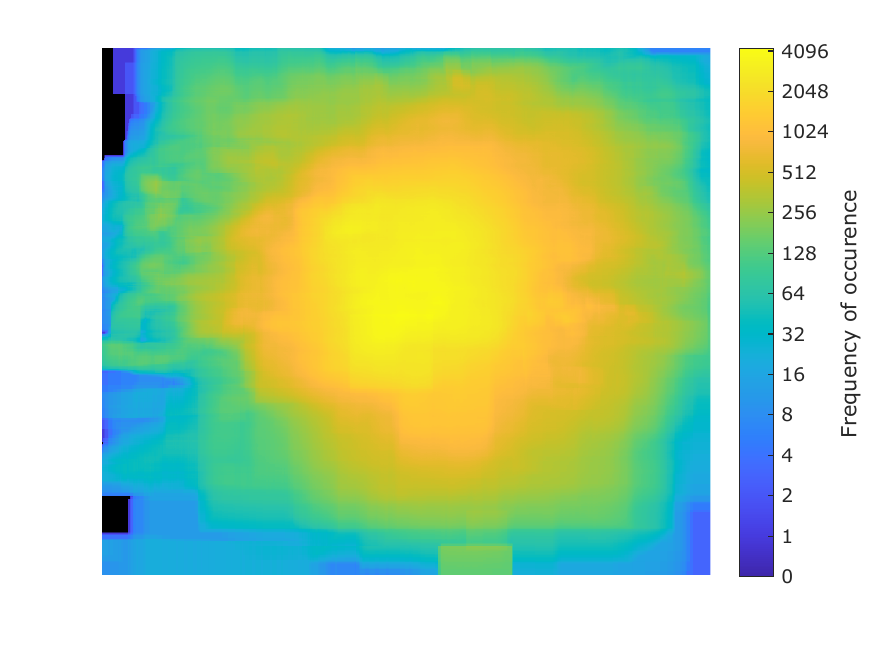}
	\caption{Annotation heatmap of the drones recorded by the Sony camera.}
	\label{fig:heatmap_color}
\end{figure}

Figures \ref{fig:heatmap_infrared} and \ref{fig:heatmap_color} show heatmaps of the average positions of annotated drones, used for training, on the individual pixels of the infrared and color images separately. Black pixels in the heatmaps represent positions that are not reached by any drone. As can be seen in Fig.\,\ref{fig:heatmap_infrared}, the average position of drones is more prominent around the center of the images compared to the boundaries particularly due to the hand held FLIR camera following the drone comfortably. The occurance of drones near the boundaries can be attributed to the heavy InfraTec camera which is difficult to steer at the required speed. The smallest bounding box has a size of 128 pixels while the largest bounding box has a size of 230886 pixels. The average bounding box size is approximately 7472 pixels. 

The color images have the drone appearing more around the center of the image compared to the boundaries. Similar to the FLIR camera the Sony $\alpha$6000 is also easy to maneuver and follow the drone. The smallest bounding box has a size of 44 pixels while the largest bounding box has a size of 368954 pixels. The average bounding box size is approximately 9794 pixels. All of the above sizes are calculated with respect to a reference image size of $1280 \times 1280$ pixels which is the default input dimension to the YOLO models used in the later section.

\section{Deep Learning Based Drone Detection}
\label{sec:DroneDetection}

After an initial application of multiple CNN architectures from the YOLO family \cite{yolov5}, \cite{yolov6}, \cite{yolov7} and one based on EfficientDet \cite{efficientDet}, it was concluded that YOLOv7 was the best performing model, particularly in terms of inference speed. The YOLOv7 model is pre-trained on the COCO dataset and the pre-trained weights are used to initialize during training. Similar to previous YOLO models, YOLOv7's architecture can be divided into a backbone, neck, and head structure.

\begin{figure*}[htbp]
	\centering
	\includegraphics[width=2\columnwidth]{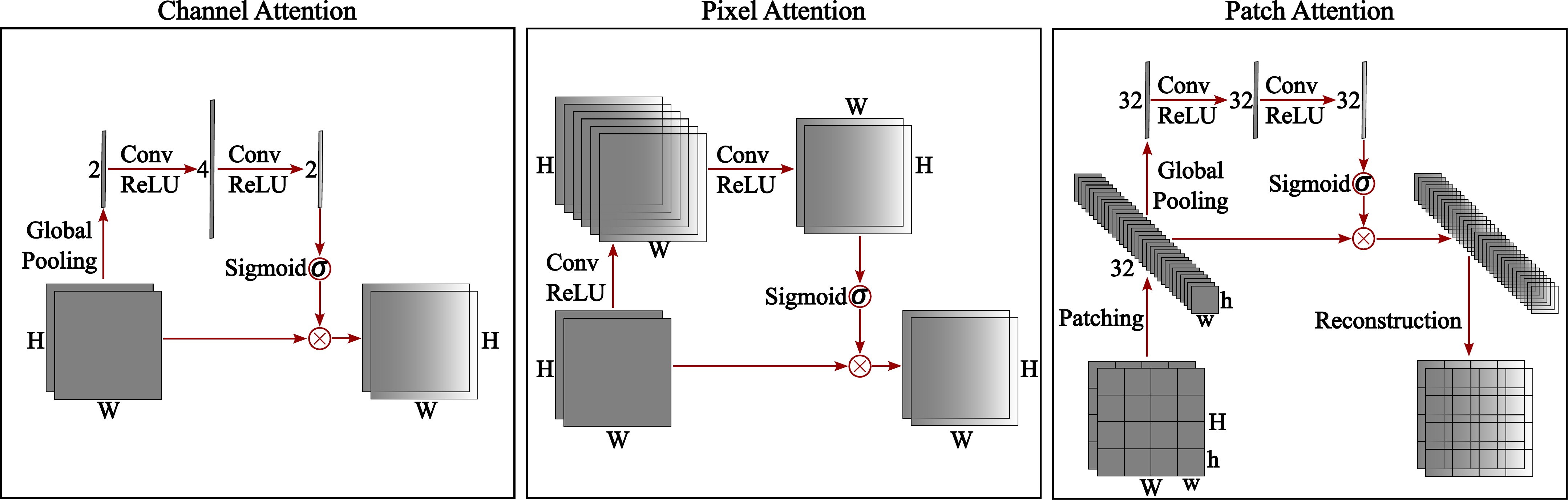}
	\caption{A simple illustration of the different attention modules.}
	\label{fig:attentionmodules}
\end{figure*} 

Apart from the basis YOLOv7 architecture, a set of simple modules and a couple of transformer modules are also experimented with, to see if an improvement in detection accuracy is significant enough to compromise with increased computation. The additional modules are illustrated in Fig.\,\ref{fig:attentionmodules}. The first module is a channel-attention (CAT) module which is used in many computer vision tasks \cite{bhattacharya2022deep, Woo2018ECCV} where each input feature map is weighted with a scalar value or an importance score, learned during training. The pixel attention (PxAT) module \cite{bhattacharya2022deep} weighs each pixel of a feature map instead of the entire feature map, which performs better in particular computer vision tasks such as image denoising. The patch attention (PAT) module divides each feature map into patches of a certain size and learns an importance score for a region in each feature map of the layer. Such modules attempt to decorrelate the feature maps to reduce redundancy and improve their relative variation. Transformer modules are particularly popular in natural language processing and large language models \cite{Vaswani2017NIPS} due to their efficiency of pattern recognition in sequential data. Their usage in deep learning for computer vision tasks has shown significant improvements in multiple applications \cite{Dosovitskiy2021ICLR, Chen2022ECCV} as well, because they are usually better at improving the detection of more salient features or redundant structures on a feature map. The classical transformer module with multihead attention structure was initially used for such tasks. These modules with multiple heads are usually quite computationally expensive, particularly in models for vision applications. Hence, the shifted window transformer (SWIN) \cite{Liu2021CVF} which is faster than a vanilla transformer, is experimented with. All the above modules are added in pyramidal backbone at multiple stages where the feature map resolution decreases. However, fewer PAT and SWIN transformer modules are used compared to CAT and PxAT due to their higher parametric complexity and memory overhead. The subset of the entire dataset for infrared and color images is divided into training and validation sets. The corresponding information is provided in Table \ref{tab:datadivision}. About $22\,\%$ of the infrared and $19\,\%$ of the color dataset are used for validation. The division of the images among the train and validation dataset is fairly homogeneous and images from both locations are present in training and validation datasets. However, certain backgrounds in the training dataset are not present in the validation dataset and vice versa.

The initial experiments with the different architectures are performed on the color image datasets and the results are tabulated in Table \ref{tab:map_archs} which shows the average performance in terms of mean average precision ($\text{MAP}_\text{IOU=0.5}$ and $\text{MAP}_\text{IOU=0.5:0.95}$) values at different intersection over union (IOU) thresholds and average processing time expressed in milliseconds. It can be seen from the results that channel and pixel attention modules improve the average precision values with a minor increment in inference time. The patch attention module increases the inference time much more while the SWIN module results in a significant increment of the inference time. A reduction in number of layers or heads can decrease the inference time for the transformer module but the precision results are not better than the simpler attention modules. Indeed, based on detection requirement, the base model might suffice in certain cases, particularly in conjunction with a good tracking algorithm. However, moving forward, the channel attention module is used since it is deemed to have a marginally better overall performance.

\begin{table}[bt]
	\centering
	\footnotesize
	\caption{Number of images split into training and validation data for FLIR Scion OTM366 and InfraTec VarioCAM HD Z.}
	\vspace{2mm}
	\label{tab:datadivision}
	\begin{tabular}{|L{2.5cm}|Q|C{2cm}|Q|C{2cm}|} \hhline{-~-~-}
		\vspace{-0.3cm}\xrowht{0.4cm}\centering\multirow{2}{*}{\begin{tabular}[c]{@{}c@{}}Number of\\ images\end{tabular}} & \hspace{0.08cm}
		& \vspace{-0.3cm}\xrowht{0.4cm}\multirow{2}{*}{Infrared} & \hspace{0.08cm}
		& \vspace{-0.3cm}\xrowht{0.4cm}\multirow{2}{*}{Color}
		\\ \hhline{~~~~~}
		\xrowht{0.3cm} & 
		& \xrowht{0.3cm} &
		& \xrowht{0.3cm}
		\\ \hhline{=~=~=}
		\xrowht{0.3cm}Training & 
		& \xrowht{0.3cm}44,300 &
		& \xrowht{0.3cm}30,440
		\\ \hhline{-~-~-}
		\xrowht{0.3cm}Validation & 
		& \xrowht{0.3cm}12,483 &		
		& \xrowht{0.3cm}7,068
		\\ \hhline{-~-~-}
	\end{tabular}
\end{table}

\begin{table*}[t]
	\centering
	\footnotesize
	\caption{MAP results of different object detection architectures for color images.}
	\vspace{2mm}
	\label{tab:map_archs}
	\begin{tabular}{|C{2.5cm}|C{2.5cm}|C{2.5cm}|C{2.5cm}|} \hhline{----}
		\vspace{-0.3cm}\xrowht{0.4cm}\multirow{2}{*}{Architecture}
		& \vspace{-0.3cm}\xrowht{0.4cm}\multirow{2}{*}{MAP$_{0.5:0.95}$}
		& \vspace{-0.3cm}\xrowht{0.4cm}\multirow{2}{*}{MAP$_{0.5}$}
		& \vspace{-0.3cm}\xrowht{0.4cm}\multirow{2}{*}{\begin{tabular}[c]{@{}c@{}}Avg. Time / Image\\ in ms\end{tabular}}
		\\
		\xrowht{0.3cm}
		& \xrowht{0.3cm}
		& \xrowht{0.3cm}
		& \xrowht{0.3cm}
		\\ \hhline{====}
		\xrowht{0.3cm}{Basis}
		& \xrowht{0.3cm} 0.679
		& \xrowht{0.3cm} 0.949
		& \xrowht{0.3cm} \textbf{19.7}
		\\ \hhline{====}
		\xrowht{0.3cm}{CAT}
		& \xrowht{0.3cm} \textbf{0.712}
		& \xrowht{0.3cm} 0.952
		& \xrowht{0.3cm} 21.3
		\\ \hhline{====}
		\xrowht{0.3cm}{PxAT}
		& \xrowht{0.3cm} 0.697
		& \xrowht{0.3cm} \textbf{0.954}
		& \xrowht{0.3cm} 22.7
		\\ \hhline{====}
		\xrowht{0.3cm}{PAT} 
		& \xrowht{0.3cm} 0.699
		& \xrowht{0.3cm} \textbf{0.954}
		& \xrowht{0.3cm} 28.1
		\\ \hhline{====}
		\xrowht{0.3cm}{SWIN-T} 
		& \xrowht{0.3cm} 0.698
		& \xrowht{0.3cm} 0.953
		& \xrowht{0.3cm} 67
		\\ \hhline{----}
	\end{tabular}
\end{table*}

\begin{table*}[t]
	\centering
	\footnotesize
	\caption{MAP results of different object detection models for infrared and color image datasets.}
	\vspace{2mm}
	\label{tab:map}
	\begin{tabular}{|C{2.5cm}|Q|C{2.5cm}|Q|C{2.5cm}|C{2.5cm}|Q|C{2.5cm}|} \hhline{-~-~--~-}
		\vspace{-0.3cm}\xrowht{0.4cm}\multirow{2}{*}{Model}
		& \hspace{0.08cm}
		& \vspace{-0.3cm}\xrowht{0.4cm}\centering{\multirow{2}{*}{Dataset}}
		& \hspace{0.08cm}
		& \vspace{-0.3cm}\xrowht{0.4cm}\multirow{2}{*}{MAP$_{0.5:0.95}$}
		& \vspace{-0.3cm}\xrowht{0.4cm}\multirow{2}{*}{MAP$_{0.5}$}
		& \hspace{0.08cm}
		& \vspace{-0.3cm}\xrowht{0.4cm}\multirow{2}{*}{\begin{tabular}[c]{@{}c@{}}Avg. Time / Image\\ in ms\end{tabular}}
		\\
		\xrowht{0.3cm}
		& 
		& \xrowht{0.3cm}
		&
		& \xrowht{0.3cm}
		& \xrowht{0.3cm}
		&
		& \xrowht{0.3cm}
		\\ 
		\hhline{=~=~==~=}
		\vspace{-0.4cm}\xrowht{0.3cm}\multirow{3}{*}{YOLOv7W6 (CAT)}
		& 
		& \xrowht{0.3cm} {\begin{tabular}[c]{@{}c@{}}Infrared\\ (1280$\times$1280)\end{tabular}}
		& 
		& \xrowht{0.3cm} 0.757
		& \xrowht{0.3cm} 0.976
		&
		& \xrowht{0.3cm} 21.2
		\\ 
		\hhline{~~-~--~-}
		& 
		& \xrowht{0.3cm} {\begin{tabular}[c]{@{}c@{}}Color\\ (1280$\times$1280)\end{tabular}}
		& 
		& \xrowht{0.3cm} 0.712
		& \xrowht{0.3cm} 0.952
		&
		& \xrowht{0.3cm} 21.3
		\\ \hhline{-~-~--~-}
	\end{tabular}
\end{table*}

Separate models are trained with infrared images and color images. The models are initialized with pre-trained weights and the default hyperparameters are used, except for the learning rates and the weight decay values which are adjusted. To train with the YOLO models, the VOC annotations are converted to text annotations. Each model is trained for 100 epochs with a batch size of 8 on an Nvidia RTX8000 workstation GPU. Table \ref{tab:map} shows the results on the validation datasets for the infrared and color datasets. It can be observed from the results that the precision metrics are relatively good and the processing is fast enough for live detection. However, while using the models on videos or live streams, detection drops between frames occur, particularly during sudden changes in environment or contrast.

\begin{figure*}
	\centering
	\begin{subfigure}[b]{0.48\textwidth}
		\centering
		\includegraphics[scale=0.5]{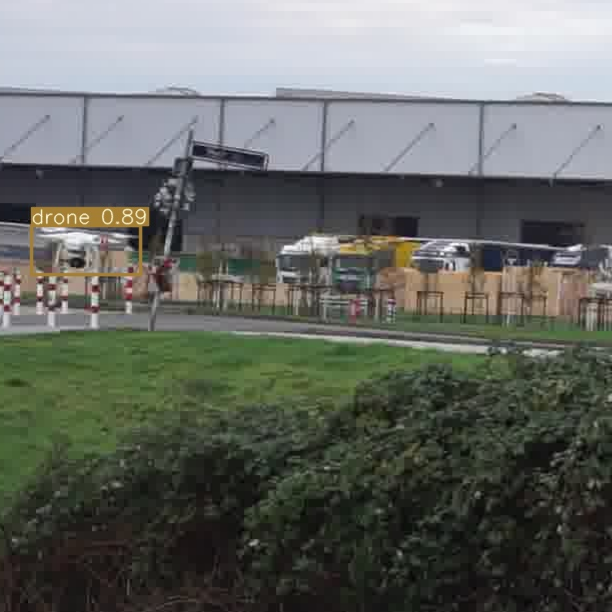}
		\caption{Detection-Sony}
		\label{fig:Yolov7example1color}
	\end{subfigure}
	\hfill
	\begin{subfigure}[b]{0.48\textwidth}
		\centering
		\includegraphics[scale=0.5]{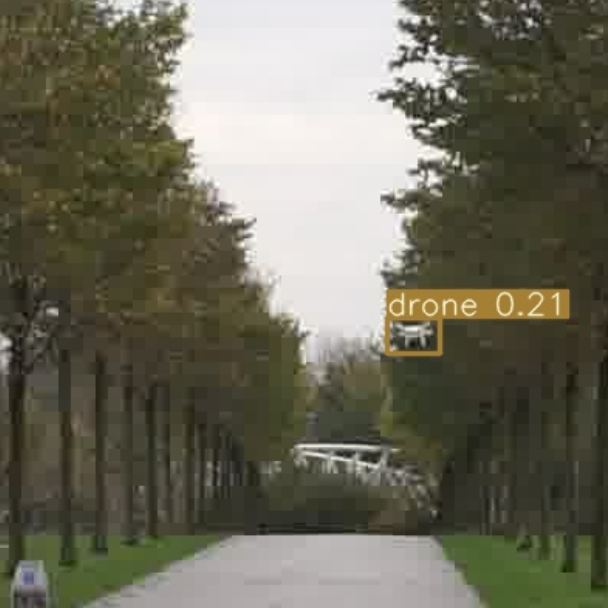}
		\caption{Detection-Sony}
		\label{fig:Yolov7example2color}
	\end{subfigure} \\ \vspace{0.25cm}
	\begin{subfigure}[b]{0.48\textwidth}
		\centering
		\includegraphics[scale=0.5]{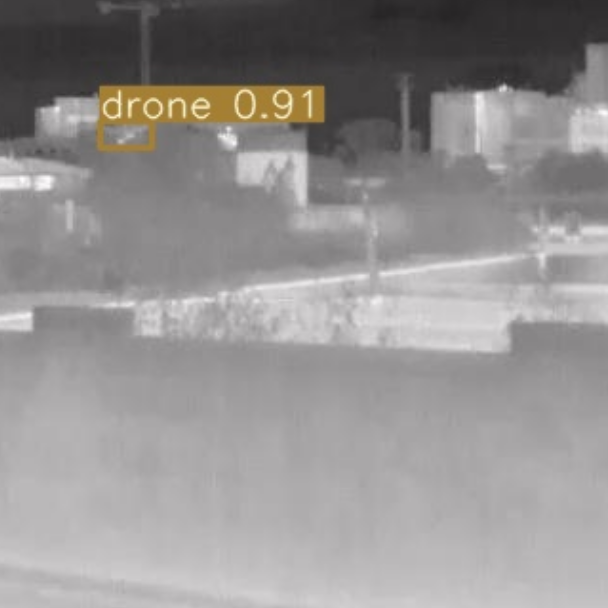}
		\caption{Detection-FLIR}
		\label{fig:Yolov7example1infrared}
	\end{subfigure}
	\hfill
	\begin{subfigure}[b]{0.48\textwidth}
		\centering
		\includegraphics[scale=0.5]{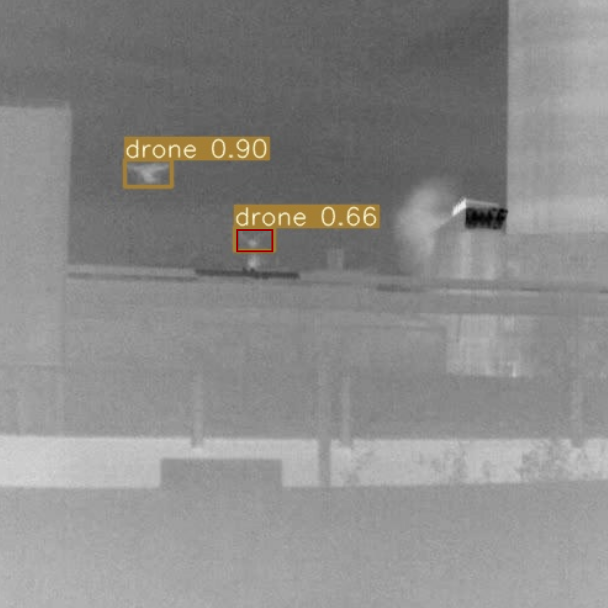}
		\caption{Detection-InfraTec}
		\label{fig:Yolov7example2infrared}
	\end{subfigure}
	\caption{Example of drone detection on some cropped sections of individual images from test data by the YOLOv7 model. Cropped sections of the entire image are shown for better visibility. The red box indicates a false positive.}
	\label{fig:exampleResults}
\end{figure*}

A selection of detection examples is shown in Fig.\,\ref{fig:exampleResults}. The yellow boxes with the confidence scores denote detection. The performance of YOLOv7 on an individual image is mostly reliable but a few exceptions do occur. On some frames, a drop in confidence score can result in a false negative when the threshold is high, as shown in Fig.\,\ref{fig:Yolov7example2color}. On the other hand, a false positive with a high confidence score can also occur as shown with the red box in Fig.\,\ref{fig:Yolov7example2infrared}, particularly in an infrared image having similar structures to a drone.

\subsection{Rule-based Tracking Method}

Although YOLOv7 performs relatively well in terms of successful detection, there are certain frames in between where the drone detection fails either due to a sudden drop of the confidence score or some degree of occlusion. In order to improve the detection drop due to a sudden change in confidence, a simple rule or condition based method is proposed. The method can be described in the following steps:

\begin{itemize}
	\item A high ($\text{conf}_h$) and a low confidence threshold ($\text{conf}_l$) are selected.
	\item The bounding boxes above $\text{conf}_l$ are preserved while only the bounding boxes above $\text{conf}_h$ are considered as valid. The method waits for the first valid detection or detections.
	\item The first valid detection or detections are used as reference. These bounding boxes and slightly larger regions around these boxes are selected for use in a cross-correlation function. This frame becomes the previous frame.
	\item The bounding boxes from the new or current frame are initially registered to the previous valid detections and the corresponding ones (bounding box or boxes based on proximity, IOU, and size) are selected.
	\item If new bounding boxes of the current frame are found and their confidence scores are above $\text{conf}_h$, no other step is required.
	\item If the confidence scores of the new boxes are lower than $\text{conf}_h$ but higher than $\text{conf}_l$, then bounding boxes are also predicted. This prediction is done with a cross-correlation method from the corresponding bounding boxes in the previous frame and the region around that boxes cropped from the new frame. Based on the estimated shift, the predicted bounding boxes for the current frame are generated and intersection over union (IOU) values with the new bounding boxes are calculated. If the IOU values are above a certain threshold, the new bounding boxes are considered as valid and their confidence scores are replaced by the previous high confidence score.
	\item If the IOU values calculated in the previous step are lower than a threshold, then the predicted bounding boxes can be considered as valid if the cross-correlation values are higher than a given threshold.
	\item If the above does not happen, the objects are either occluded or not detected. The confidence score of any corresponding bounding box is replaced with 0. The process waits for the next valid detections.
\end{itemize} 

\begin{figure*}[htbp]
	\centering
	\includegraphics[width=2\columnwidth]{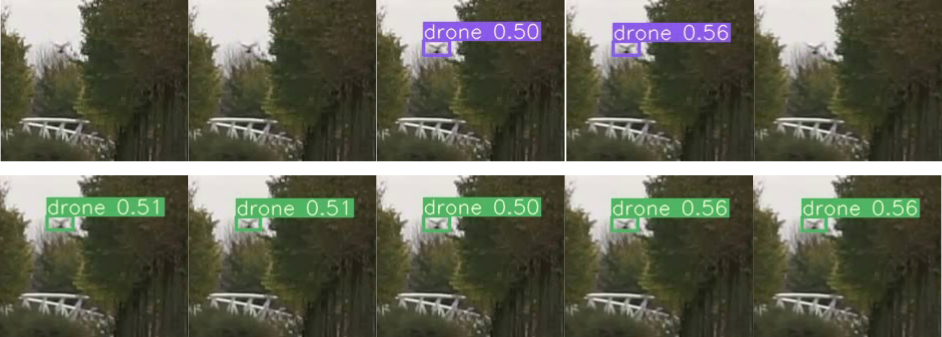}
	\caption{Consecutive frames from a test video having missing detections with YOLOv7 (top) and the same frames with successful detections after the additional method (bottom).}
	\label{fig:Trackingvsnotracking}
\end{figure*} 

In the above method, the different threshold values can be experimented with. The proposed method running on top of YOLOv7 successfully detects more drones compared to the usage of only YOLOv7 and reduces detection drops. The amount of improvement depends on the choice of different thresholds and usually ranges from $2$ to $10$\,\% more drones detected, depending on the video. Figure \ref{fig:Trackingvsnotracking} shows an example of the method which makes additional detections compared to only YOLOv7. However, there are still many instances where detection is unsuccessful due to very low confidence scores. While the system is dependent on YOLO making an initial detection per frame, any confidence score below the low threshold ($\text{conf}_l$) will result in a discontinuation of tracking. The method is based on one prior frame and can be improved by the inclusion of multiple prior frames for a more accurate prediction of the drone's position and movement.
\begin{figure}[htbp]
	\centering
	\includegraphics[width=0.98\columnwidth]{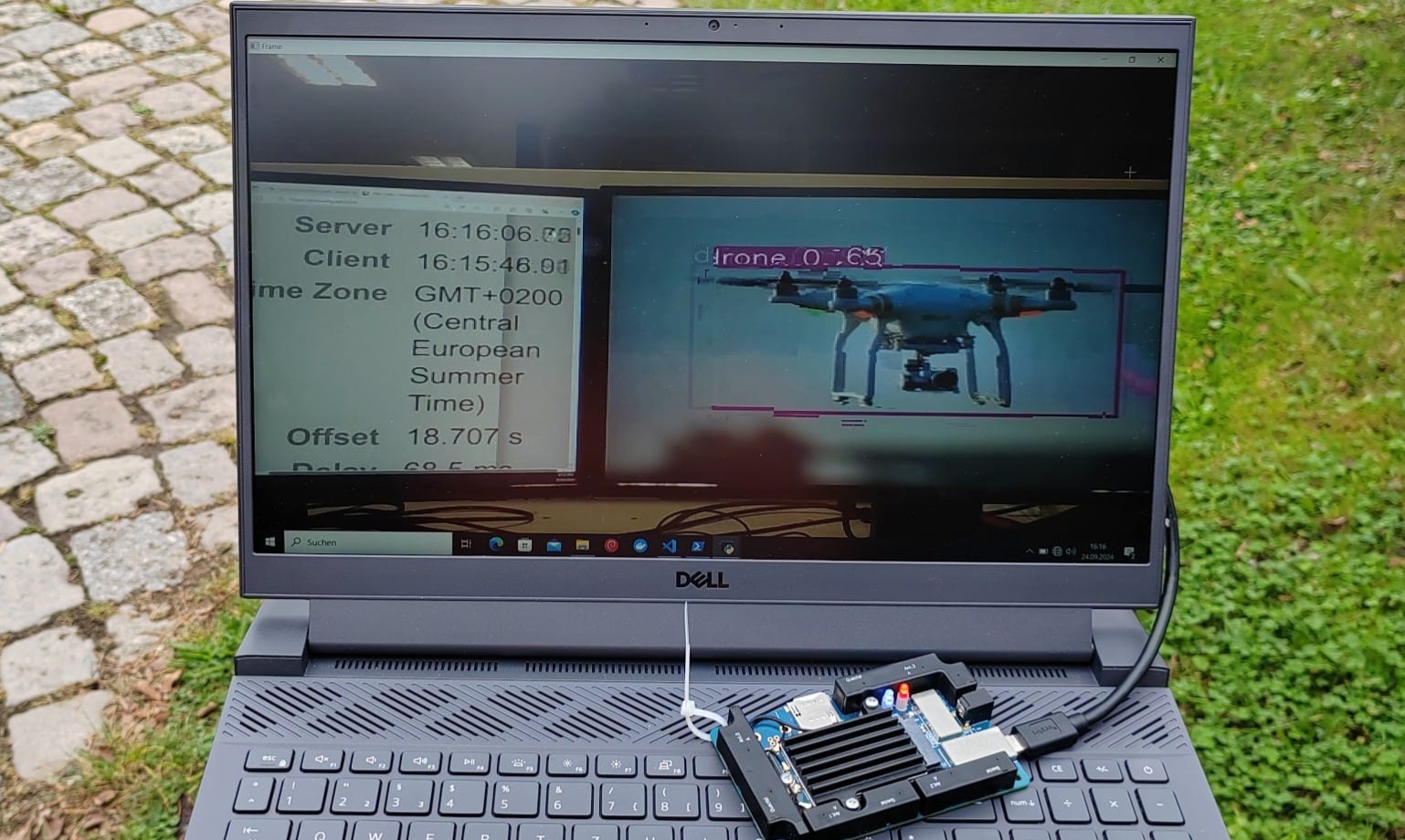}
	\caption{A simple test for live drone detection.}
	\label{fig:livedetection}
\end{figure}

A simple and fast test is performed where an indoor computer runs a video of a flying drone and the displayed video on a monitor is recorded by a USB camera placed at a certain distance. The recording is simultaneously transmitted from the indoor network to the outdoor network and is received by a laptop connected to the 5G network and running the containerized model. The model detects the drone and displays it on the screen, as shown in Fig.\,\ref{fig:livedetection}. This test is done for multiple times during a day. On the laptop, a $15$ fps frame rate is achieved without any visible stutter. An approximate latency of $350\,\text{ms}$ is observed which is primarily contributed by the USB camera, the parameter settings in GStreamer pipeline (jitter buffer), and the detection model running on the laptop with a Nvidia RTX3050 processing unit. However, a proper test setup needs to be built with a real use-case scenario and accurate measurement setup.

\section{Conclusion}
\label{sec:Conclusion}
This work is a continuation of our previous work related to a drone detection system based on YOLOv7. The data required for this task is extended to include a annotated color image and a multi-channel image dataset. The overall dataset is described and analyzed, and the infrared and color image datasets are used to train multiple CNN models based on YOLOv7. Their evaluation shows a reasonably reliable behavior when it comes to drone detection, with occasional false negatives due to a drop in confidence or a false positive, particularly in infrared images. A simple rule or condition based method is proposed to reduce the detection drops and improve tracking. The multi-channel dataset is proposed, with the idea that it should result in an improved detection performance compared to a single image source. Since the dataset is generated from a pre-trained YOLO model, its performance should not be evaluated by retraining with the same model. As a future work, the datasets should be trained with a different model and the results should be compared. Given, the challenges of image registration across multiple image types, it might be appropriate to use a camera system with multiple sensors which are accurately calibrated, aligned and synchronized. For the live drone detection system, an experimental setup should be constructed with a real scenario and accurate measurement setup.

\section{Acknowledgement}
\label{sec:Acknowledgement}
This work was part of the ``Digital Sensor-2-Cloud Campus Platform'' (DS2CCP) project, which is funded by the Federal Ministry of Defense under the dtec.bw program.

\end{document}